\documentclass[journal,twoside,web]{ieeecolor}
\usepackage{tmech}
\usepackage{cite}
\usepackage[switch,pagewise]{lineno}
\usepackage{url,subfigure}
\usepackage{booktabs}
\usepackage{multirow}
\usepackage{threeparttable}
\usepackage{amsmath}
\usepackage{bm}
\usepackage{balance}
\usepackage{amsfonts}
\usepackage{scalerel}
\usepackage{tikz}
\usetikzlibrary{svg.path}
\definecolor{orcidlogocol}{HTML}{A6CE39}
\tikzset{
  orcidlogo/.pic={
    \fill[orcidlogocol] svg{M256,128c0,70.7-57.3,128-128,128C57.3,256,0,198.7,0,128C0,57.3,57.3,0,128,0C198.7,0,256,57.3,256,128z};
    \fill[white] svg{M86.3,186.2H70.9V79.1h15.4v48.4V186.2z}
                 svg{M108.9,79.1h41.6c39.6,0,57,28.3,57,53.6c0,27.5-21.5,53.6-56.8,53.6h-41.8V79.1z M124.3,172.4h24.5c34.9,0,42.9-26.5,42.9-39.7c0-21.5-13.7-39.7-43.7-39.7h-23.7V172.4z}
                 svg{M88.7,56.8c0,5.5-4.5,10.1-10.1,10.1c-5.6,0-10.1-4.6-10.1-10.1c0-5.6,4.5-10.1,10.1-10.1C84.2,46.7,88.7,51.3,88.7,56.8z};
  }
}
\newcommand\orcidicon[1]{\href{https://orcid.org/#1}{\mbox{\scalerel*{
\begin{tikzpicture}[yscale=-1,transform shape]
\pic{orcidlogo};
\end{tikzpicture}
}{|}}}}
\usepackage{hyperref}
\hypersetup{
    colorlinks=true,
    linkcolor=blue,
    filecolor=black,      
    urlcolor=black,
    citecolor=purple
}
\usepackage{tikz-network}

\newcommand{\etal}{\textit{et al. }}

\def\BibTeX{{\rm B\kern-.05em{\sc i\kern-.025em b}\kern-.08em
		T\kern-.1667em\lower.7ex\hbox{E}\kern-.125emX}}
\begin{document}
	\title{\huge Freespace Optical Flow Modeling for Automated Driving}
	\author{
		Yi Feng$^{\orcidicon{0009-0005-4885-0850}\,}$, 
		Ruge Zhang$^{\orcidicon{0009-0008-7526-8980}\,}$, 
		Jiayuan Du$^{\orcidicon{0000-0003-1589-9111}\,}$,
		Qijun Chen$^{\orcidicon{0000-0001-5644-1188}\,}$,
		and Rui Fan$^{\orcidicon{0000-0003-2593-6596}\,}$
		\thanks{
			This research was supported by the National Key R\&D Program of China under Grant 2020AAA0108100, the National Natural Science Foundation of China under Grant 62233013, the Science and Technology Commission of Shanghai Municipal under Grant 22511104500, and the Fundamental Research Funds for the Central Universities.
			 \textit{(Yi Feng and Ruge Zhang share first authorship.)} \textit{(Corresponding author: Rui Fan)}}
		\thanks{Yi Feng, Jiayuan Du, Qijun Chen, and Rui Fan are with the College of Electronics \& Information Engineering, Shanghai Research Institute for Intelligent Autonomous Systems, the State Key Laboratory of Intelligent Autonomous Systems, and Frontiers Science Center for Intelligent Autonomous Systems, Tongji University, Shanghai 201804, P. R. China (e-mails: fengyi@ieee.org, \{dujiayuan, qjchen\}@tongji.edu.cn, rui.fan@ieee.org).}
		\thanks{Ruge Zhang is with the High-Performance Computer Research Center, Institute of Computing Technology, Chinese Academy of Sciences, Beijing 100190, P R. China (e-mail: zhangruge23@mails.ucas.ac.cn).}
	}
	\maketitle
	
	\def\dt{\Delta t}
	\def\dphi{\Delta\varphi}
	
\begin{abstract}
Optical flow and disparity are two informative visual features for autonomous driving perception. They have been used for a variety of applications, such as obstacle and lane detection. The concept of ``U-V-Disparity'' has been widely explored in the literature, while its counterpart in optical flow has received relatively little attention. Traditional motion analysis algorithms estimate optical flow by matching correspondences between two successive video frames, which limits the full utilization of environmental information and geometric constraints. Therefore, we propose a novel strategy to model optical flow in the collision-free space (also referred to as drivable area or simply freespace) for intelligent vehicles, with the full utilization of geometry information in a 3D driving environment. We provide explicit representations of optical flow and deduce the quadratic relationship between the optical flow component and the vertical coordinate. Through extensive experiments on several public datasets, we demonstrate the high accuracy and robustness of our model. Additionally, our proposed freespace optical flow model boasts a diverse array of applications within the realm of automated driving, providing a geometric constraint in freespace detection, vehicle localization, and more. We have made our source code publicly available at \url{https://mias.group/FSOF}.
\end{abstract}
	
	\begin{IEEEkeywords}
		Optical flow, autonomous driving perception, freespace, automated driving, vehicle localization.
	\end{IEEEkeywords}
	
\section{Introduction}
\label{sec.introduction}	
\IEEEPARstart{W}{ith} the advancements in sensor technology and navigation positioning systems, the advanced driver assistance system (ADAS) has emerged as a crucial component of intelligent vehicles \cite{jimenez2016advanced}. It utilizes multiple sensors to monitor the surrounding environment and provides real-time information to the front end of the automated driving system \cite{kaempchen2003data}. Simultaneously, the emergence and evolution of various computer vision algorithms have driven continuous improvements in the accuracy and performance of the environmental perception functionality in ADAS systems \cite{fan2020sne}. Motion analysis poses a fundamental and challenging problem in environmental perception \cite{wu2001optical}, which entails the estimation of 2D or 3D object motion using dynamic scene sequences captured across multiple successive video frames. Optical flow estimation focuses specifically on analyzing 2D motion, whereas scene flow estimation deals with 3D motion analysis \cite{Zhai2021Optical,vedula1999three}. Optical flow and scene flow estimations hold profound significance in establishing high-level cognitive ability in 3D scene understanding and play a pivotal role in downstream ADAS applications, such as mobile robot navigation \cite{yu2022accurate} and semantic scene parsing \cite{fan2020sne}.

\begin{figure}[!t]
		\centering
		\includegraphics[width=0.47\textwidth]{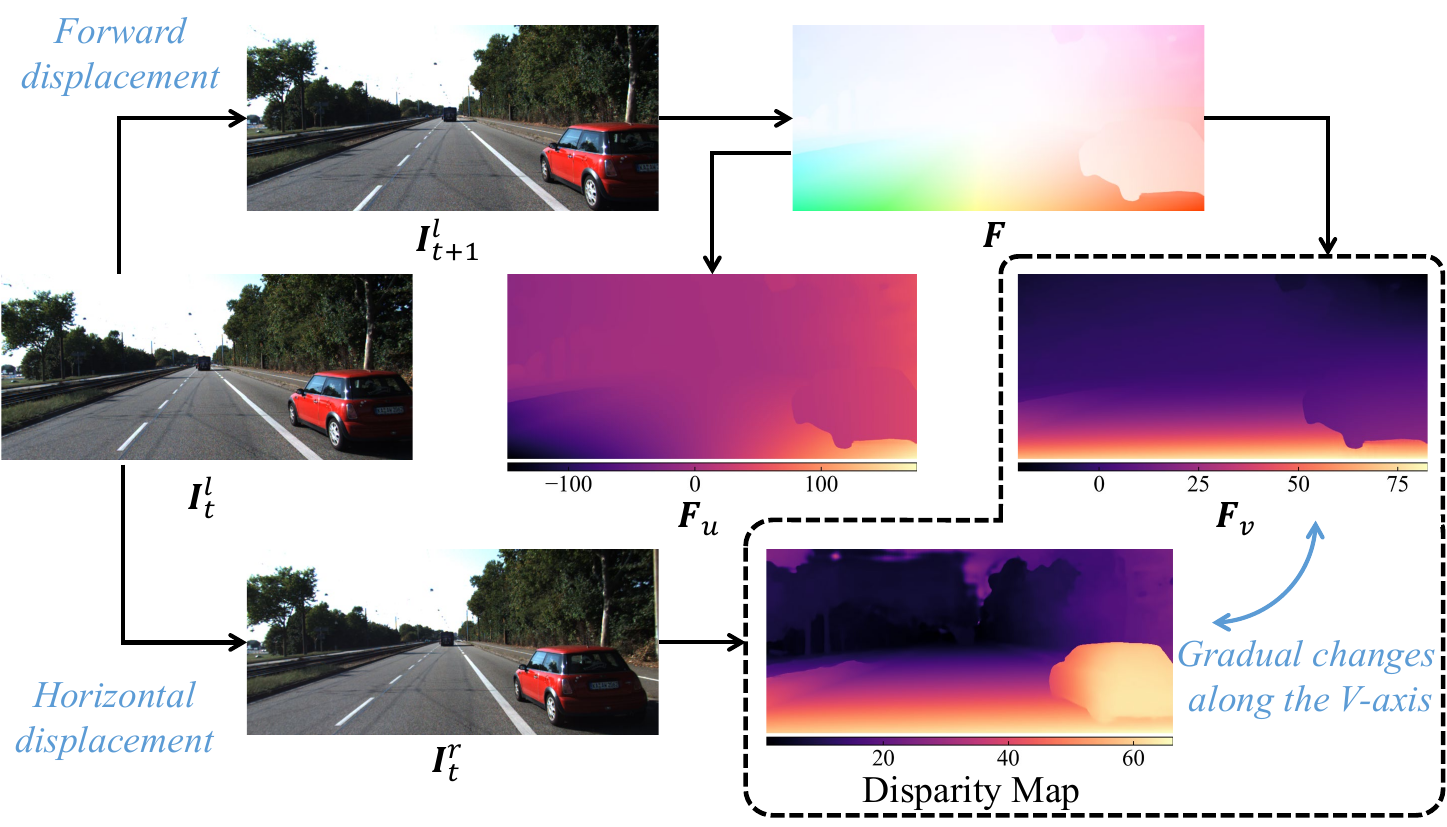}
		\caption{Difference and relationship between optical flow and disparity. $\boldsymbol{F}_u$ and $\boldsymbol{F}_v$ are the horizontal and vertical components of optical flow $\boldsymbol{F}$. }
		\label{fig.fu_fv_disp}
  \end{figure}

Depth estimation and object tracking are two crucial tasks in ADAS. Depth estimation typically involves using a stereo rig to acquire disparity information \cite{fan2018road}, which is inversely proportional to the depth. In autonomous driving scenarios, the depth map often exhibits gradual changes along the $V$-axis of the image plane due to the geometric features of the road surface \cite{fan2019pothole}. On the other hand, object tracking is commonly accomplished or aided by optical flow information, which describes the pixel-level relationship by a two-dimensional motion field based on luminosity consistency. Fig. {\ref{fig.fu_fv_disp}} illustrates the difference and relationship between optical flow and disparity in autonomous driving scenarios. $\boldsymbol{I}_t^l$ and $\boldsymbol{I}_t^r$ represent the RGB images obtained by the left and right cameras at time $t$, respectively, while $\boldsymbol{I}_{t+1}^l$ represents the image captured by the left camera at time $t+1$. The optical flow map $\boldsymbol{F}\in\mathbb{R}^{H\times W\times 2}$ is obtained by calculating the displacements of corresponding pixels between two successive video frames, where $\boldsymbol{F}_u$ and $\boldsymbol{F}_v$ store the horizontal and vertical displacements, respectively. This process introduces a 3D translation and a slight rotation from the camera's perspective. In contrast, the disparity map represents corresponding pixels that are horizontally shifted between the stereo image pair. In this process, it is considered that a monocular camera produces a horizontal translation $b$ (the stereo rig baseline) within a very short period of time. As shown in Fig. \ref{fig.fu_fv_disp}, both the disparity map and $\boldsymbol{F}_v$ exhibit similar distributions: they both change gradually along the $V$-axis.

Hu \etal \cite{Zhencheng2005Complete} conducted a comprehensive analysis of the relationship between disparity and pixel coordinates, which can be expressed as follows:
\begin{equation}
		\begin{aligned}
			\left\{
			\begin{array}{l} 
				u=\Phi(x^w,\Delta) + u_0 \\ 
				v=\Psi(z^w,\Delta) + v_0 \\
			\end{array}
			\right.,
		\end{aligned}
		\label{eq.UVDisp}
	\end{equation}
where $\boldsymbol{p}^w=[x^w,y^w,z^w]^\top$ represents a 3D point in the world coordinate system, $\boldsymbol{p}=[u,v]^\top$ denotes its projection in the left image, $\boldsymbol{p}_0=[u_0,v_0]^\top$ represents the coordinates of the principal point, $\Phi$ and $\Psi$ are two linear functions, and $\Delta$ denotes the disparity of $\boldsymbol{p}$. Additionally, the geometric characteristics in 3D space can be abstracted into plane equations with various expressions. By combining these plane equations with (\ref{eq.UVDisp}), disparity maps can be mapped to the \textit{U-Disparity} or \textit{V-Disparity} domains using cumulative voting algorithms \cite{fan2021rethinking}.
	
Our work is motivated by two observations. First, $\boldsymbol{F}_v$ exhibits a similar distribution to that of the disparity map. Second, the physical meanings of optical flow and disparity are highly similar, as both describe the coordinate offset of corresponding pixels between two successive video frames. Inspired by the \textit{U-V-Disparity} work mentioned above, we aim to explore the regularities in the distribution of freespace optical flow through mathematical modeling of the 3D driving environment and the integration of camera parameters and odometry information.
		
This article is structured as follows: Section \ref{sec.preliminaries} discusses two types of optical flow definitions and relevant literature. In Section \ref{sec.methodology}, we derive the mathematical relationship between an arbitrary on-road pixel $\boldsymbol{p}$ and its optical flow $\boldsymbol{F}(\boldsymbol{p})$ for two different types of optical flow data formats. In Section \ref{sec.implementation_experiments}, we present the experimental results concerning model validation and robustness evaluation. Section  \ref{sec.discussion} provides a detailed discussion of various applications based on our proposed model. Finally, we summarize our work in Section \ref{sec.conclusion}.
	
\section{Literature Review}
\label{sec.preliminaries}
	
Optical flow can be defined in two different ways. Traditional optical flow algorithms consider optical flow as the partial derivative of image intensity with respect to time \cite{Tu2019survey}. This definition characterizes optical flow as a distribution of apparent velocity of brightness patterns in the current image \cite{Horn1981Determining, Lucas1981Iterative}. However, this definition assumes small inter-frame displacements and short time intervals, which may not always hold true in real-world scenarios. On the other hand, data-driven optical flow estimation methods rely on large and accurate optical flow datasets to enhance the performance of neural networks \cite{mayer2016large, wang2021cot, shi2023flowformer++, mehl2023spring}. These datasets define optical flow ground truth as the displacement vector of corresponding pixels between two adjacent frames. This is because that the instantaneous velocity of pixels cannot be directly measured, and calibration errors are inevitable \cite{Geiger2012Are}.
	
In accordance with the two aforementioned definitions, optical flow approaches can be categorized as prior knowledge-driven or data-driven. The former ones, exemplified by Horn and Schunck's flow estimation framework \cite{Horn1981Determining}, aim to minimize distortions in optical flow and prefer solutions that exhibit smoothness. Nevertheless, these approaches assume smoothness in optical flow across the entire image and the invariance of pixel intensity between frames, making them difficult to apply in real-world scenarios. Following Horn's work, Lucas and Kanade \cite{Lucas1981Iterative} introduced a local invariance constraint to estimate sparse optical flow. However, when the input data is noisy, this method performs poorly due to its reliance on the least squares criterion. To address illumination variation problems, Brox \etal \cite{brox2004high} introduced a gradient constancy assumption. However, this assumption is only effective for linear illumination changes and is not capable of handling large vehicle displacements.

Data-driven algorithms, such as FlowNet \cite{Dosovitskiy_2015_ICCV}, employ deep neural networks and large optical flow datasets to enhance the accuracy of optical flow estimation. These methods yield superior results compared to traditional approaches. However, they may not fully exploit the environment geometry information, relying heavily on hardware capabilities and the quality of optical flow datasets.
	
Our work builds upon the two definitions of optical flow and aims to systematically construct freespace optical flow models with geometry constraints. By deriving the explicit relationship between an on-road pixel $\boldsymbol{p}$ and its optical flow $\boldsymbol{F}(\boldsymbol{p})$, we effectively address the problems mentioned above.
	
\section{Methodology}
\label{sec.methodology}
	
\begin{figure}[!t]
		\centering
		\includegraphics[width=0.25\textwidth]{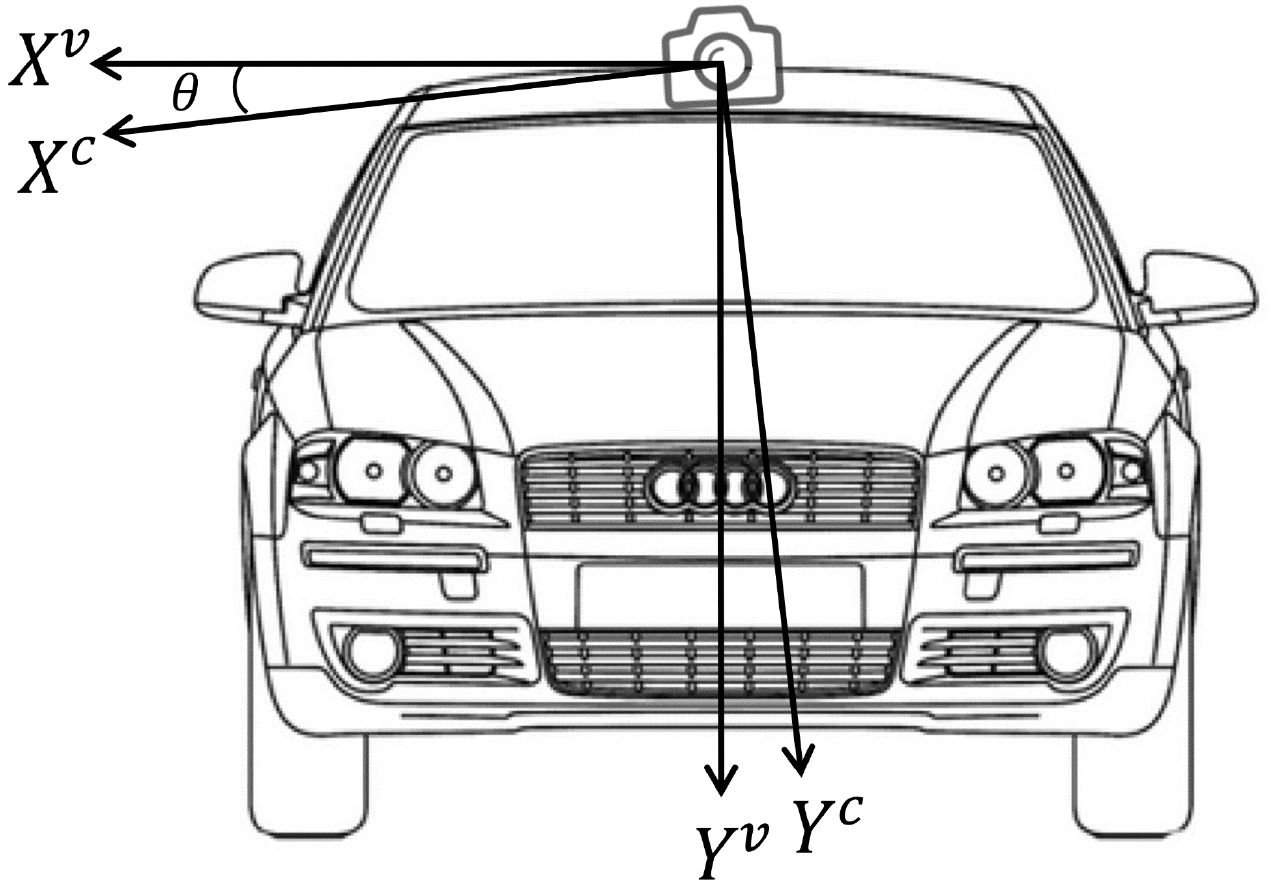}
		\label{fig.illus1}
		\subfigure[]{
			\includegraphics[width=0.22\textwidth]{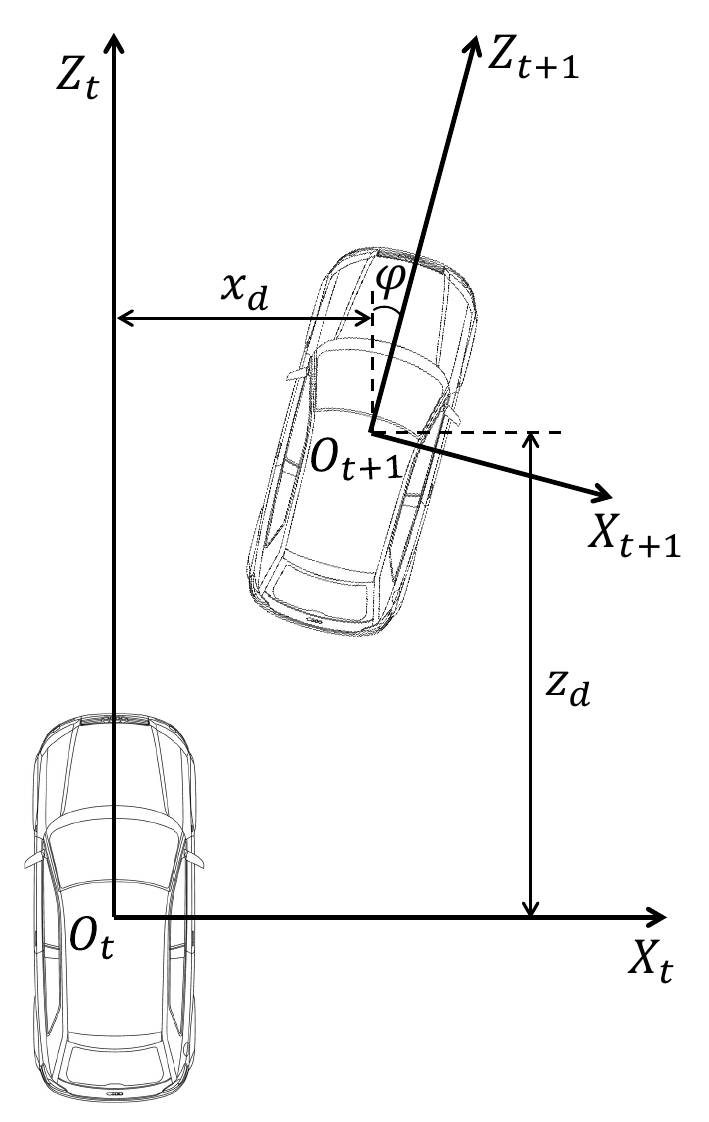}
			\label{fig.disp_model}
		}
		\subfigure[]{
			\includegraphics[width=0.22\textwidth]{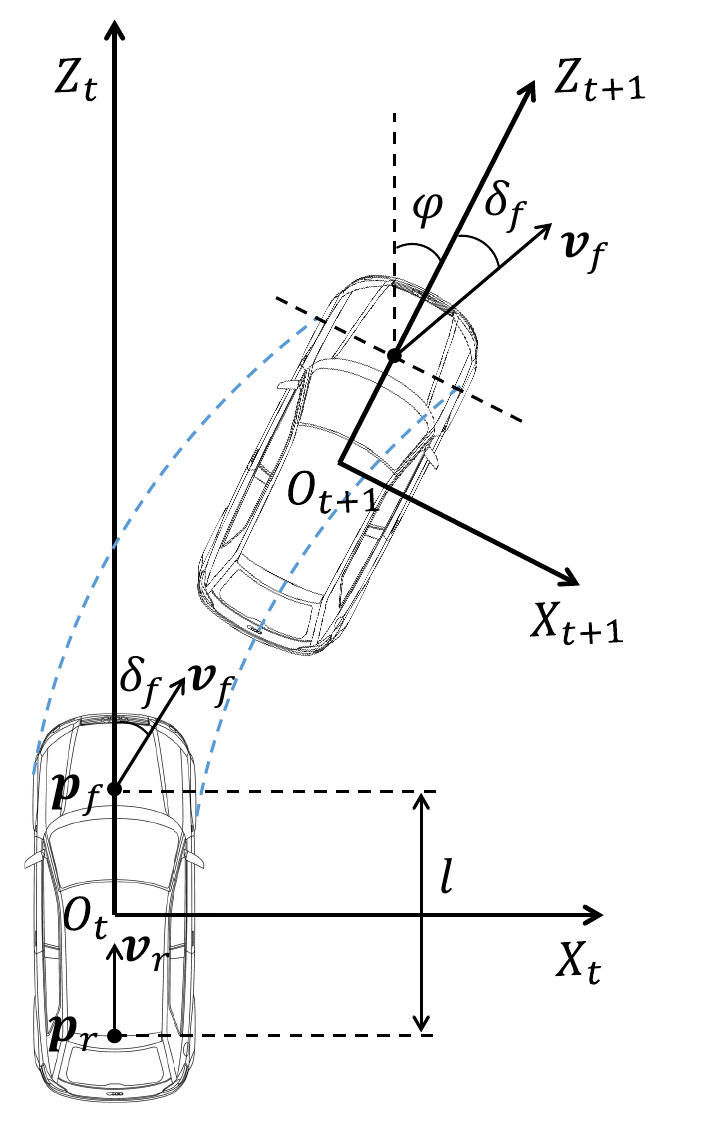}
			\label{fig.vel_model}
		}
		
		\caption{An illustration of the two models: (a) displacement-based model; (b) velocity-based model.}
		\label{fig.model}
	\end{figure}

An example of the monocular system mounted on autonomous vehicles for environmental perception is illustrated in Fig. \ref{fig.model}. We assume that the monocular system fixed above the vehicle is a basic pinhole camera model with the camera intrinsic matrix $\boldsymbol{K}$, and the vehicle with the monocular camera can be considered as a rigid body.
	
In our model, three coordinates are considered: vehicle coordinate system (VCS), camera coordinate system (CCS), and pixel coordinate system (PCS). Due to the rigid connection between the monocular camera and the vehicle body, we place the CCS origin at the same position as the VCS origin, with the $Y$-axis pointing vertically downward and the $Z$-axis parallel to the vehicle’s motion direction. Taking into account the potential mounting error and centrifugal force experienced when a car turns, there exists a roll angle $\theta$ along the $Z$-axis between the VCS and CCS. Therefore, a 3D point $\boldsymbol{p}{_t^c}=[x{_t^c},y{_t^c},z{_t^c}]^\top$ in the CCS at time $t$ can be linked to a 3D point $\boldsymbol{p}{^v_t}=[x{^v_t},y{^v_t},z{^v_t}]^\top$ in the VCS using $\boldsymbol{p}{_t^c}=\boldsymbol{R}_{\theta}\boldsymbol{p}{^v_t}$, where 
	\begin{equation}
		\begin{aligned}
			\boldsymbol{R}_{\theta} =\begin{bmatrix}
				\cos\theta &  \sin\theta & 0  \\
				-\sin\theta & \cos\theta & 0 \\
				0 & 0 & 1 \\
			\end{bmatrix}
		\end{aligned}
		\label{eq.r_theta}
	\end{equation}
is the rotation matrix. Moreover, we consider the road surface as a horizontal plane, because the unevenness of the road and jolts while driving are negligible compared to the scale of the camera's field of view. Therefore, the road surface can be described by a plane equation: $y^v = h$, where $h$ is the mounting height of the camera from the ground. 
	
Our model establishes the relationship between optical flow and pixel coordinates, explicitly representing the optical flow with respect to vehicle poses and camera parameters. As discussed in Section \ref{sec.preliminaries}, there are two distinct definitions of optical flow. Therefore, we derive two different forms of the freespace optical flow model: the displacement-based freespace optical flow model, represented by $\boldsymbol{f}_d$, and the velocity-based freespace optical flow model, represented by $\boldsymbol{f}_v$. The choice between the two models should be made based on the optical flow data format. The displacement-based model is applicable when the optical flow is encoded in inter-frame displacement, whereas the velocity-based model is applicable when the optical flow is encoded in pixel apparent velocity.
	
\subsection{Displacement-Based Freespace Optical Flow Modeling}
As mentioned in Section \ref{sec.preliminaries}, optical flow is commonly defined as the displacement between two adjacent frames in most public datasets, such as Sintel \cite{butler2012naturalistic}, Flying Chairs \cite{Dosovitskiy_2015_ICCV}, KITTI \cite{Geiger2012Are, Menze2015Object}, and Middlebury \cite{Baker2010Database}. Therefore, we first establish a displacement-based freespace optical flow model, which explicitly describes the relationship between a freespace point $\boldsymbol{p}$ in the PCS and its optical flow $\boldsymbol{f}_d$.
	
	For an arbitrary 3D point $\boldsymbol{p}{^v_t}=[x{^v_t},y{^v_t},z{^v_t}]^\top$ in the VCS, it can be projected to $\boldsymbol{p}{_t}=[u{_t},v{_t}]^\top$ in PCS of the video frame captured at time $t$ as follows:
	\begin{equation}
		\begin{aligned}
			z{^c_t}\tilde{\boldsymbol{p}}{_t}=\boldsymbol{K}\boldsymbol{R}_{\theta}\boldsymbol{p}{^v_t},
		\end{aligned}
		\label{eq.frame1}
	\end{equation}
	where $z{^c_t}= z{^v_t}$ is the depth of point $\boldsymbol{p}$ in the CCS of the frame at time $t$, and $\tilde{\boldsymbol{p}}{_t}$ represents the homogeneous coordinates of $\boldsymbol{p}{_t}$. Combining road surface equation ($y^v=h$) with (\ref{eq.frame1}), we can calculate $\boldsymbol{p}{^v_t}$ with respect to $\boldsymbol{p}{_t}$:
	\begin{equation}
		\begin{aligned}
			\boldsymbol{p}{^v_t}
			=h\begin{bmatrix}
				\displaystyle{\frac{f_y(u{_t}-u_0)\cos\theta-f_x(v{_t}-v_0)\sin\theta}{f_y(u{_t}-u_0)\sin\theta+f_x(v{_t}-v_0)\cos\theta}} \vspace{1ex} \\
				1  \\
				\displaystyle{\frac{f_xf_y}{f_y(u{_t}-u_0)\sin\theta+f_x(v{_t}-v_0)\cos\theta}}\\
			\end{bmatrix},
		\end{aligned}
		\label{eq.Wxyz}
	\end{equation}
	where $f_x$ and $f_y$ are the horizontal and vertical camera focal lengths, respectively. In order to simplify (\ref{eq.Wxyz}), we define $\lambda_1$ and $\lambda_2$ as follows:
	\begin{equation}
		\begin{aligned}
			\left\{
			\begin{array}{l}
				\displaystyle{\lambda_1 = \frac{u{_t}-u_0}{f_x}\sin\theta+\frac{v{_t}-v_0}{f_y}\cos\theta}, \vspace{1ex}\\ 
				\displaystyle{\lambda_2 = \frac{u{_t}-u_0}{f_x}\cos\theta-\frac{v{_t}-v_0}{f_y}\sin\theta}. \\
			\end{array}
			\right.
		\end{aligned}
		\label{eq.lambda12}
	\end{equation}
	Therefore, (\ref{eq.Wxyz}) can be rewritten as follows:
	\begin{equation}
		\begin{aligned}
			\boldsymbol{p}{^v_t}=\displaystyle{\frac{h}{\lambda_1}}\begin{bmatrix}
				\lambda_2  \\
				\lambda_1  \\
				1  \\
			\end{bmatrix}.
		\end{aligned}
		\label{eq.p_vt}
	\end{equation}	
Since the camera mounted on the vehicle typically has a fixed distance to the ground plane, only the yaw angle $\varphi$ (rotation around the $Y$-axis) and a translation $\boldsymbol{d}=[x_d,0,z_d]^\top$ in the $X$ and $Z$ directions, respectively, should be considered. Therefore, we can obtain the following relationship:
	\begin{equation}
		\boldsymbol{p}{^c_{t+1}}
		=\boldsymbol{R}_{\theta}\boldsymbol{R}_{\varphi}(\boldsymbol{p}{^v_t}-\boldsymbol{d}),
	\end{equation}
	where 
	\begin{equation}
		\boldsymbol{R}_{\varphi}=
		\begin{bmatrix}
			\cos\varphi  &0  &-\sin\varphi \\
			0  &1  &0\\
			\sin\varphi  &0  &\cos\varphi\\
		\end{bmatrix}
	\end{equation}
 represents the rotation with respect to $\varphi$. Therefore, the perspective projection for the second frame is as follows:
	\begin{equation}
		\begin{aligned}
			z^c_{t+1}\tilde{\boldsymbol{p}}_{t+1}=\boldsymbol{K}\boldsymbol{p}{^c_{t+1}}
			=\boldsymbol{K}\boldsymbol{R}_{\theta}\boldsymbol{R}_{\varphi}(\boldsymbol{p}{^v_t}-\boldsymbol{d}),
		\end{aligned}
		\label{eq.frame2}
	\end{equation}
	where $z^c_{t+1}=(x{^v_t}-x_d)\sin\varphi+(z{^v_t}-z_d)\cos\varphi$ is the depth of point $\boldsymbol{p}{^c_{t+1}}$. Combining (\ref{eq.frame1}) and (\ref{eq.frame2}) yields:
	\begin{equation}
		\begin{aligned}
			\begin{bmatrix}
                    \boldsymbol{f}_d\\
				0 \\
			\end{bmatrix}=\boldsymbol{K}\boldsymbol{R}_{\theta}(\frac{1}{z_{t+1}}\boldsymbol{R}_{\varphi}(\boldsymbol{p}{^v_t}-\boldsymbol{d})-\frac{1}{z_t}\boldsymbol{p}{^v_t}).
		\end{aligned}
		\label{eq.fufv0}
	\end{equation}
 By plugging (\ref{eq.p_vt}) into (\ref{eq.fufv0}), we deduce the displacement-based freespace optical flow $\boldsymbol{f}_d$ as follows:
		\begin{equation}
			\begin{aligned}
				\boldsymbol{f}_d
				&=\boldsymbol{M}
				\begin{bmatrix}
					\displaystyle{\frac{\lambda_3\cos\varphi-\lambda_4\sin\varphi}{\lambda_3\sin\varphi+\lambda_4\cos\varphi}-\lambda_2} \vspace{1ex}\\ 
					\displaystyle{\frac{\lambda_1h}{\lambda_3\sin\varphi+\lambda_4\cos\varphi}-\lambda_1} \\
				\end{bmatrix},
			\end{aligned}
			\label{eq.f_final}
		\end{equation}
		where 
		\begin{equation}
			\begin{aligned}
                        \boldsymbol{M} &=\begin{bmatrix}
						f_x\cos\theta  &f_x\sin\theta \\
						-f_y\sin\theta &f_y\cos\theta\\
					\end{bmatrix},\\
					\lambda_3 &= \lambda_2h-\lambda_1x_d, \\
					\lambda_4 &= h - \lambda_1z_d.
			\end{aligned}
			\label{eq.lambda34}
		\end{equation}
		Therefore, we derive the explicit relationship between the optical flow and the pixel coordinates incorporating odometry and camera parameters.
		
		\subsection{Velocity-Based Optical Flow Modeling}

    As mentioned in Section \ref{sec.preliminaries}, the optical flow has also been defined as the (apparent or instantaneous) velocities of movement of brightness patterns in an image. Therefore, we also develop a velocity-based freespace optical flow model to accommodate such data format.

    To derive the velocity-based freespace optical flow model, we introduce the Ackermann steering geometry \cite{wong2022theory} for four-wheel autonomous vehicle kinematics modeling. As illustrated in Fig. {\ref{fig.vel_model}}, $\delta_f$ denotes the steering angle of the front wheel, $\boldsymbol{p}_r=[x_r,z_r]^\top$ and $\boldsymbol{p}_f=[x_f,z_f]^\top$ respectively represent the coordinates of vehicle's rear and front axle centers, $\boldsymbol{v}_r=[\dot{x}_r, \dot{z}_r]^\top$ and $\boldsymbol{v}_f=[\dot{x_f}, \dot{z_f}]^\top$ respectively denote the velocities of the rear and front axles ($||\boldsymbol{v}_r||_2=||\boldsymbol{v}_f||_2=v_r$), $l$ denotes the distance between the front and rear axles, and $\varphi$ represents the vehicle's turning angle. With no sideslip of the front and rear axles, the kinematics constraints can be expressed as follows:
		\begin{equation}
			\begin{split}
				&\dot{x}_f\cos(\varphi+\delta_f) = \dot{z}_f\sin(\varphi+\delta_f), \\
				&\dot{x}_r\cos\varphi = \dot{z}_r\sin\varphi,
			\end{split}
			\label{eq.sideslipConst}
		\end{equation}
		where 
		\begin{equation}
			\begin{split}
				&\dot{x}_r = v_r\sin\varphi, \ \ \ \dot{z}_r = v_r\cos\varphi.
			\end{split}
			\label{eq.speedConst}
		\end{equation}
		According to the positions of the front and rear axle centers, we can obtain the following expressions:
		\begin{equation}
			\begin{split}
				x_f &= x_r + l\sin\varphi, \ \ \ z_f = z_r + l\cos\varphi.
			\end{split}
			\label{eq.posConst}
		\end{equation}   
		Plugging (\ref{eq.speedConst}) and (\ref{eq.posConst}) into (\ref{eq.sideslipConst}) results in the Ackermann kinematic model as follows:
		\begin{equation}
			\begin{aligned}
				\left\{
				\begin{array}{l}
					\displaystyle{\dot{\varphi} = v_r\frac{\tan\delta_f}{l}}, \vspace{0.5ex}\\
					\dot{x}_r = v_r\sin\varphi, \vspace{0.5ex}\\ 
					\dot{z}_r = v_r\cos\varphi.\\
				\end{array}
				\right.
			\end{aligned}
		\end{equation}
            To derive the velocity-based freespace optical flow model $\boldsymbol{f}_v$, we assume that the time interval $\Delta t$ between two adjacent video frames approaches zero: 
		\begin{equation}
				\boldsymbol{f}_v
				=
				\lim_{\Delta t\to 0}\frac{\boldsymbol{f}_d}{\Delta t}
				=
				\lim_{\Delta t\to 0} \boldsymbol{M}\boldsymbol{q},
			\label{eq.fufvVel}
		\end{equation}
            where 
            \begin{equation}
				\boldsymbol{q}=
				\begin{bmatrix}
     \displaystyle{
     \frac{\lambda_1(\lambda_2z_d-x_d)\cos\dphi-(\lambda_4+\lambda_2\lambda_3)\sin\dphi}{(\lambda_3\sin\dphi+\lambda_4\cos\dphi)/\dt} 
     }
     \vspace{1ex}\\ 
     \displaystyle{
     \lambda_1\frac{h(1-\cos\dphi)-\lambda_3\sin\dphi+\lambda_1z_d\cos\dphi}{(\lambda_3\sin\dphi+\lambda_4\cos\dphi)/\dt}
    }
				\end{bmatrix}.
			\label{eq.fufvVel2}
		\end{equation}
		According to the limit theorem, we have:
		\begin{equation}
			\begin{aligned}
				&\lim_{\dt\to 0}\frac{\Delta x_d}{\dt} = \dot{x}_r = v_r\sin\varphi = 0, \\ 
				&\lim_{\dt\to 0}\frac{\Delta z_d}{\dt} = \dot{z}_r = v_r\cos\varphi = v_r, \\
				&\lim_{\dt\to 0}\frac{\sin\dphi}{\dt} = \lim_{\dt\to 0}\cos\dphi\dot{\varphi} = v_r\frac{\tan\delta_f}{l}, \\ 
				&\lim_{\dt\to 0}\cos\dphi = 1. \\ 
			\end{aligned}
			\label{eq.conn}
		\end{equation}
		Plugging (\ref{eq.conn}) into (\ref{eq.fufvVel}) results in:
		\begin{equation}
			\begin{aligned}
				\boldsymbol{f}_v
				&=
				\frac{v_r}{h}\boldsymbol{M}
				\begin{bmatrix}
					\displaystyle{\lambda_1\lambda_2-h(1+\lambda{_2^2})\frac{\tan\delta_f}{l}} \\ 
					\displaystyle{\lambda{_1^2}-\lambda_1\lambda_2h\frac{\tan\delta_f}{l}} \\
				\end{bmatrix}.
			\end{aligned}
			\label{eq.F}
		\end{equation}
        (\ref{eq.f_final}) and (\ref{eq.F}) establish the fundamental relationships between pixel coordinates $\boldsymbol{p}$ and their optical flow $\boldsymbol{F}(\boldsymbol{p})$ in a general monocular camera system. These relationships demonstrate that optical flow can be explicitly derived from camera model parameters and vehicle poses, eliminating the need for the brightness constancy assumption \cite{Horn1981Determining}, which can be challenging to satisfy and may lead to estimation errors. With accurate vehicle poses and camera parameters, our proposed freespace optical flow model can be effectively deployed in various downstream autonomous driving applications.
		
	\subsection{Simplified Optical Flow Models for Special Cases}
 
        In most driving scenarios, the vehicle moves primarily along the $Z$-axis with minimal rotation and little bias in the $X$ direction. As a result, the yaw angle $\varphi$ and the offset $x_d$ between two adjacent frames can be approximated to zero. Consequently, (\ref{eq.f_final}) can be simplified as follows:
		\begin{equation}
				\boldsymbol{f}_d
				=\frac{\lambda_1}{\frac{h}{z_d}-\lambda_1}
				\begin{bmatrix}
					u{_t} - u_0\\
					v{_t} - v_0\\
				\end{bmatrix}.
			\label{eq.fufvNophi}
		\end{equation}
            Furthermore, given that the steering angle $\delta_f$ is negligible, (\ref{eq.F}) can be simplified as follows:
		\begin{equation}
				\boldsymbol{f}_v
				=\frac{v_r\lambda_1}{h}
				\begin{bmatrix}
					u{_t} - u_0\\
					v{_t} - v_0
				\end{bmatrix}.
		\end{equation}
		Moreover, when there are no centrifugal effects during straight-line motion and the camera's mounting error (the nonzero roll angle) is negligible, the optical flow can be expressed in its simplest form as follows:
		\begin{equation}
			\begin{aligned}
				\begin{array}{l}
					\displaystyle{\boldsymbol{f}_d=\frac{z_d}{\frac{hf_y}{v{_t}-v_0}-z_d}
					\begin{bmatrix}
						u{_t} - u_0\\
						v{_t} - v_0\\
					\end{bmatrix}},
                        \vspace{1ex}\\
					\displaystyle{\boldsymbol{f}_v=\frac{v_r(v{_t}-v_0)}{hf_y}
					\begin{bmatrix}
						u{_t} - u_0\\
						v{_t} - v_0\\
					\end{bmatrix}.
                        }
				\end{array}
			\end{aligned}
			\label{eq.fdfv_simple}
		\end{equation}
		Therefore, it can be observed that the vertical optical flow component exhibits a quadratic relationship with respect to the vertical pixel coordinates in the freespace area. Leveraging this insight, we can adopt strategies similar to those used in U-V-Disparity analysis to effectively depict freespace and incorporate additional geometric constraints for other autonomous driving applications. 
  
		\section{Experiments}
		\label{sec.implementation_experiments}
    
            \begin{figure*}[!t]
			\centering
			\includegraphics[width=0.999\textwidth]{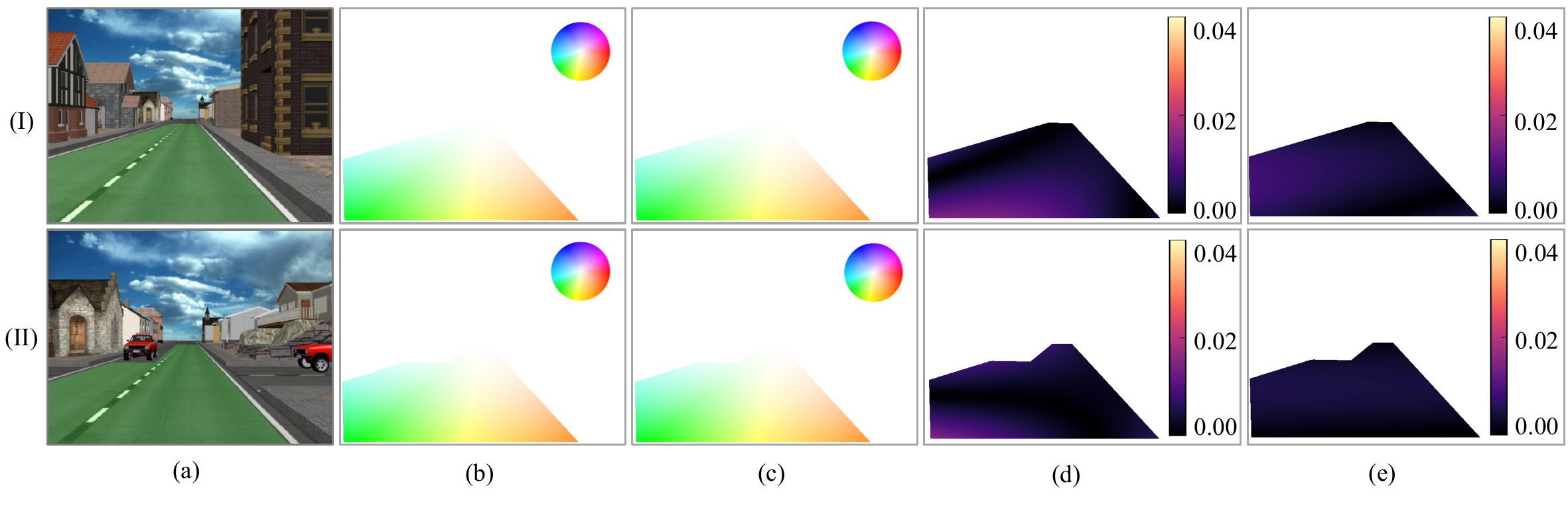}
			\caption{Qualitative experimental results on the CVC12 dataset: (a) RGB images with freespace shown in green; (b) optical flow ground truth; (c) estimated optical flow; (d) absolute error of $\boldsymbol{F}_u$ modeling; (e) absolute error of $\boldsymbol{F}_v$ modeling; (I) simple scenario; (II) complex scenario. }
			\label{fig.cvc12}
		\end{figure*}

            \begin{figure*}[!t]
			\centering
			\includegraphics[width=0.999\textwidth]{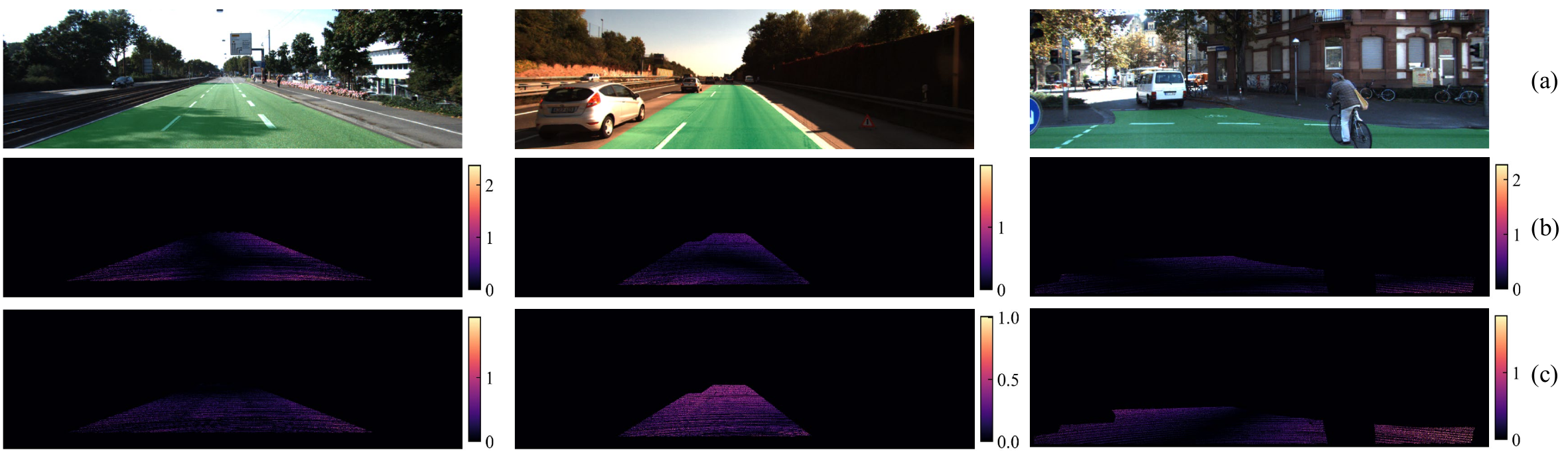}
			\caption{Qualitative experimental results on the KITTI dataset: (a) RGB images with freespace shown in green; (b) absolute error of $\boldsymbol{F}_u$ modeling; (c) absolute error of $\boldsymbol{F}_v$ modeling. }
			\label{fig.KITTI}
		\end{figure*}
  
		This section presents a comprehensive evaluation of our proposed model, including effectiveness validation, quantitative and qualitative assessments, robustness testing, and real-time performance quantification.

            As discussed in Section \ref{sec.methodology}, the optical flow can be explicitly represented by pixel coordinates and vehicle poses. This representation can be validated through surface fitting. If the optical flow modeling is incorrect, a significant fitting error may occur, which can serve as a criterion for measuring the accuracy of our model.

            \subsection{Evaluation Metrics}

           We use four evaluation metrics: 
\begin{itemize}
    \item average angular error
            \begin{equation}
				e_\text{A} = \frac{1}{N}\sum_{\boldsymbol{p}}
				\cos^{-1}\bigg(
				\frac{\langle \boldsymbol{F}(\boldsymbol{p}),\widehat{  \boldsymbol{F}}
    (\boldsymbol{p})\rangle +1 }
				 { \sqrt{ \vert\vert \boldsymbol{F}( \boldsymbol{p})\vert\vert_2^2+1)    (\vert\vert \widehat {\boldsymbol{F}}(\boldsymbol{p})\vert\vert_2^2+1) }  }
				\bigg),
			\label{eq.AAE}
		\end{equation}
  \item average end-point error
            \begin{equation}
            e_\text{E} = \frac{1}{N}\sum_{\boldsymbol{p}}\vert\vert\boldsymbol{F}(\boldsymbol{p})-\widehat{\boldsymbol{F}}(\boldsymbol{p})\vert\vert_2,
		\label{eq.AEE}
		\end{equation}
  \item and average absolute errors of $\boldsymbol{F}_u$ and $\boldsymbol{F}_v$ modeling
            \begin{equation}
            e_\text{U} = \frac{1}{N}\sum_{\boldsymbol{p}}\vert \boldsymbol{F}_{u}(\boldsymbol{p})-\widehat{\boldsymbol{F}}_{u}(\boldsymbol{p})\vert,
		\label{eq.Fu_error}
		\end{equation}
            \begin{equation}
            e_\text{V} = \frac{1}{N}\sum_{\boldsymbol{p}}\vert \boldsymbol{F}_{v}(\boldsymbol{p})-\widehat{\boldsymbol{F}}_{v}(\boldsymbol{p})\vert,
		\label{eq.FV_error}
		\end{equation}
\end{itemize}
  to quantify the effectiveness of our proposed optical flow models, where $\boldsymbol{F}$ and $\widehat{\boldsymbol{F}}$ respectively denote the ground-truth and estimated optical flow maps, $\boldsymbol{F}_{u,v}$ and $\widehat{\boldsymbol{F}}_{u,v}$ respectively represent the ground-truth and estimated horizontal and vertical optical flow maps, $\boldsymbol{p}$ denotes a pixel in the freespace area, and $N$ represents the total number of valid pixels used for model evaluation. 

                        \begin{table}[!t]
			\begin{center}
                \fontsize{7.6}{11.5}\selectfont
				\caption{Quantitative experimental results on the CVC12 dataset.}
				\label{table.cvc12}
				\setlength\arrayrulewidth{0.9pt}
				\begin{tabular}{cccc}
					\toprule
					Scenario Type
					& Vehicle Speed 
					& $e_\text{A}$ (Rad) 
					& $e_\text{E}$ (Pixel) \\
					\hline
					Simple	&$\times1$ 	&0.003 	&0.609 	 \\
					Simple	&$\times4$ 	&0.003 	&0.704 	 \\
					Complex	&$\times1$ 	&0.089 	&0.534	 \\
					Complex	&$\times4$ 	&0.030  &0.244 	 \\
					\bottomrule
				\end{tabular}
			\end{center}
		\end{table}

            \begin{figure*}[!t]
			\centering
			\includegraphics[width=0.999\textwidth]{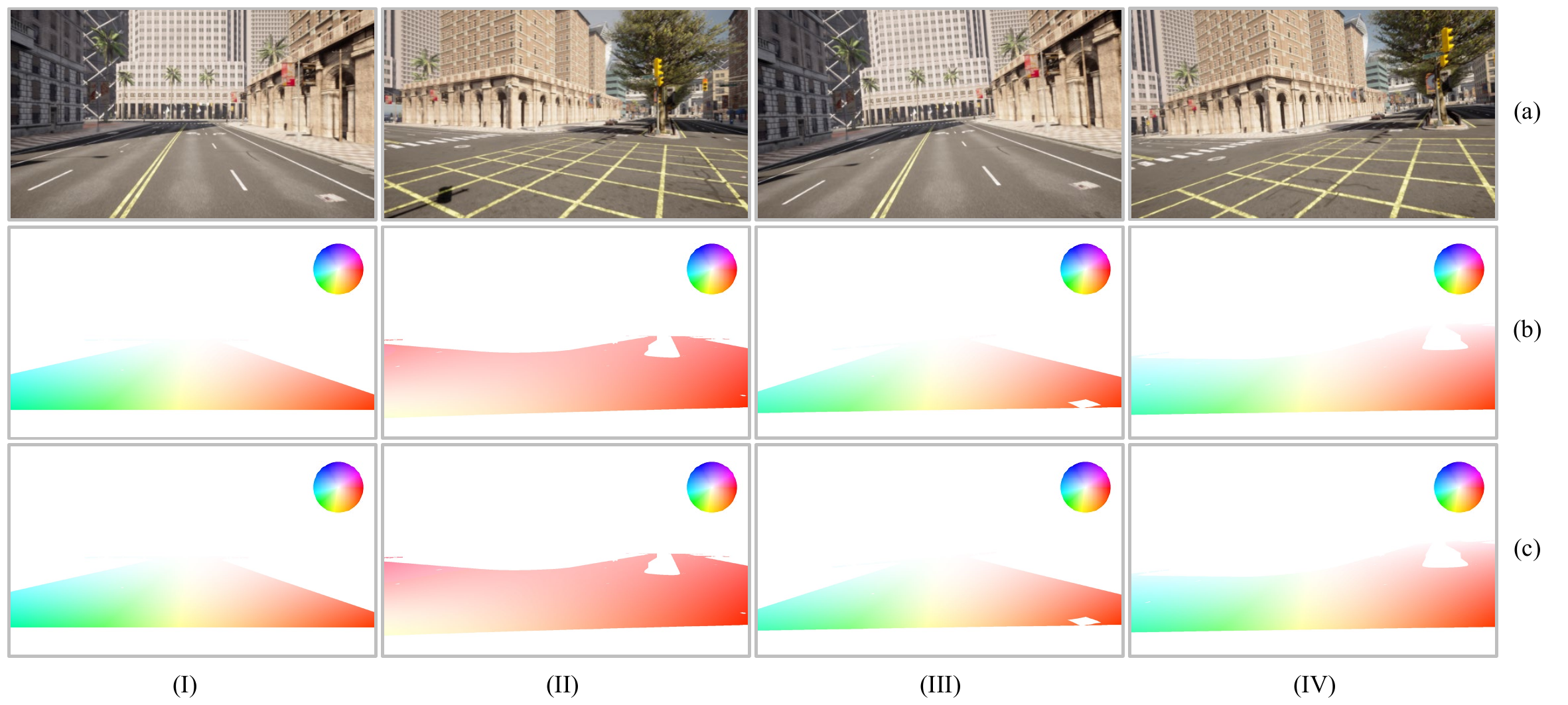}
			\caption{
   Qualitative experimental results on the CARLA dataset: (a) RGB images; (b) optical flow ground truth; (c) estimated optical flow; (I) straight-going with $\theta=0^\circ$; (II) turning with $\theta=0^\circ$; (III) straight-going  with $\theta=5^\circ$; (IV) turning with $\theta=5^\circ$.
   }
			\label{fig.carla}
		\end{figure*}

				\begin{table}[!t]
			\begin{center}
                    \fontsize{7.6}{11.5}\selectfont
				\caption{Quantitative experimental results on the KITTI dataset.}
				\label{table.kitti}
				\setlength\arrayrulewidth{0.7pt}
				\renewcommand{\arraystretch}{1.2}
				\begin{tabular}{c|cccc}
					\toprule
					{Scenario Type} 
                        &{$e_\text{A}$ (Rad)}
					&{$e_\text{E}$ (Pixel)}
					&{$e_\text{U}$ (Pixel)} 
					&{$e_\text{V}$ (Pixel)} \\
					\hline				
					Straight-going	&0.036 &0.921   &0.255 	&0.465 \\
					Turning			&0.050 &1.005	&0.599 	&0.814 \\
					\bottomrule
				\end{tabular}
			\end{center}
		\end{table}

            \subsection{Datasets}
            We utilize the following three datasets to quantify the performance of our proposed optical flow models:
            \begin{enumerate}
                \item the CVC12 optical flow \cite{onkarappa2014optical} dataset, which offers dense optical flow ground truth (in pixels) for both simple and complex synthetic driving scenarios with respect to different vehicle speeds, making it suitable for a comprehensive evaluation of our proposed displacement-based model's validity;
                \item the KITTI flow 2012 and 2015 \cite{Geiger2012Are, Menze2015Object} datasets, which provide sparse optical flow ground truth (in pixels) captured in real-world driving scenarios, making it suitable for a comprehensive evaluation of our proposed displacement-based model's robustness and effectiveness;
                \item the CARLA \cite{dosovitskiy2017carla} dataset (a synthetic dataset which we created using the CARLA simulator), which provides dense optical flow ground truth (instantaneous velocities of movement) for synthetic driving scenarios, making it suitable for a comprehensive evaluation of our proposed velocity-based model.
            \end{enumerate}
        
		\subsection{Model Effectiveness Validation}

            The qualitative results on the CVC12 dataset are shown in Fig. \ref{fig.cvc12}. It can be observed that the modeling error is minimal and its distribution is relatively uniform, validating the correctness of our proposed displacement-based model. Table \ref{table.cvc12} provides the corresponding $e_\text{A}$ and $e_\text{E}$ results, where $e_\text{A}$ is less than 0.1 rad and $e_\text{E}$ is less than one pixel. Based on these results, it is evident that our displacement-based optical flow modeling process exhibits high accuracy and robustness, regardless of varying vehicle speeds and diverse driving scenarios.
  
		\begin{table}[!t]
			\begin{center}
                \fontsize{7.6}{11.5}\selectfont
				\caption{Quantitative experimental results on the CARLA dataset.}
				\label{table.carla}
				\setlength\arrayrulewidth{0.7pt}
				\begin{threeparttable}
					\begin{tabular}{cccc}
						\hline
						\multicolumn{1}{c}{Driving Direction} 
						& \multicolumn{1}{c}{$\theta$ (Degrees)} 
						& \multicolumn{1}{c}{$e_\text{A}$ (Rad/s)} 
						& \multicolumn{1}{c}{$e_\text{E}$ (Pixel/s)}  \\

						\hline

						&\ \ 0
						&0.039
						&0.526  \\

						&\ \ 5	
						&0.038
						&0.516 \\
						
						Straight-going	
						&\ -5
						&0.040
						&0.566 \\

						&\ 10
						&0.034	
						&0.521  \\

						&-10
						&0.041
						&0.627  \\
						
						\hline
						
						&\ \ 0
						&0.034
						&1.260\\

						&\ \ 5
						&0.032
						&1.285\\
						
						Turning	
						&\ -5
						&0.035
						&1.275 \\

						&\ 10
						&0.033
						&1.357 \\

						&-10
						&0.037
						&1.325 \\

						\hline
					\end{tabular}
				\end{threeparttable}
			\end{center}
		\end{table}
		
            Furthermore, we use the KITTI flow datasets to evaluate the effectiveness and robustness of our proposed displacement-based model. The qualitative and quantitative experimental results are presented in Fig. \ref{fig.KITTI} and Table \ref{table.kitti}, respectively. It can be observed that our model achieves an $e_\text{A}$ of less than 0.05 rad and an $e_\text{E}$ of less than one pixel in almost all scenarios, demonstrating the exceptional performance of our model in real-world scenarios.
		
            Finally, we present the evaluation of our velocity-based models with respect to two types of driving directions and varying camera roll angles on the CARLA dataset. The experimental results are shown in Fig. \ref{fig.carla} and Table \ref{table.carla}. It can be observed that our model achieves high accuracy in $\theta$ estimation, with an $e_\text{A}$ of less than 0.05 rad/s and an $e_\text{E}$ of less than 1.5 pixels/s, respectively. These results strongly indicate the effectiveness of our proposed velocity-based optical flow model.

		\subsection{Optical Flow Modeling in the Special Case}
            \label{sec.optical_flow_special_cases}

\begin{figure}[!t]
			\centering
			\includegraphics[width=0.47\textwidth]{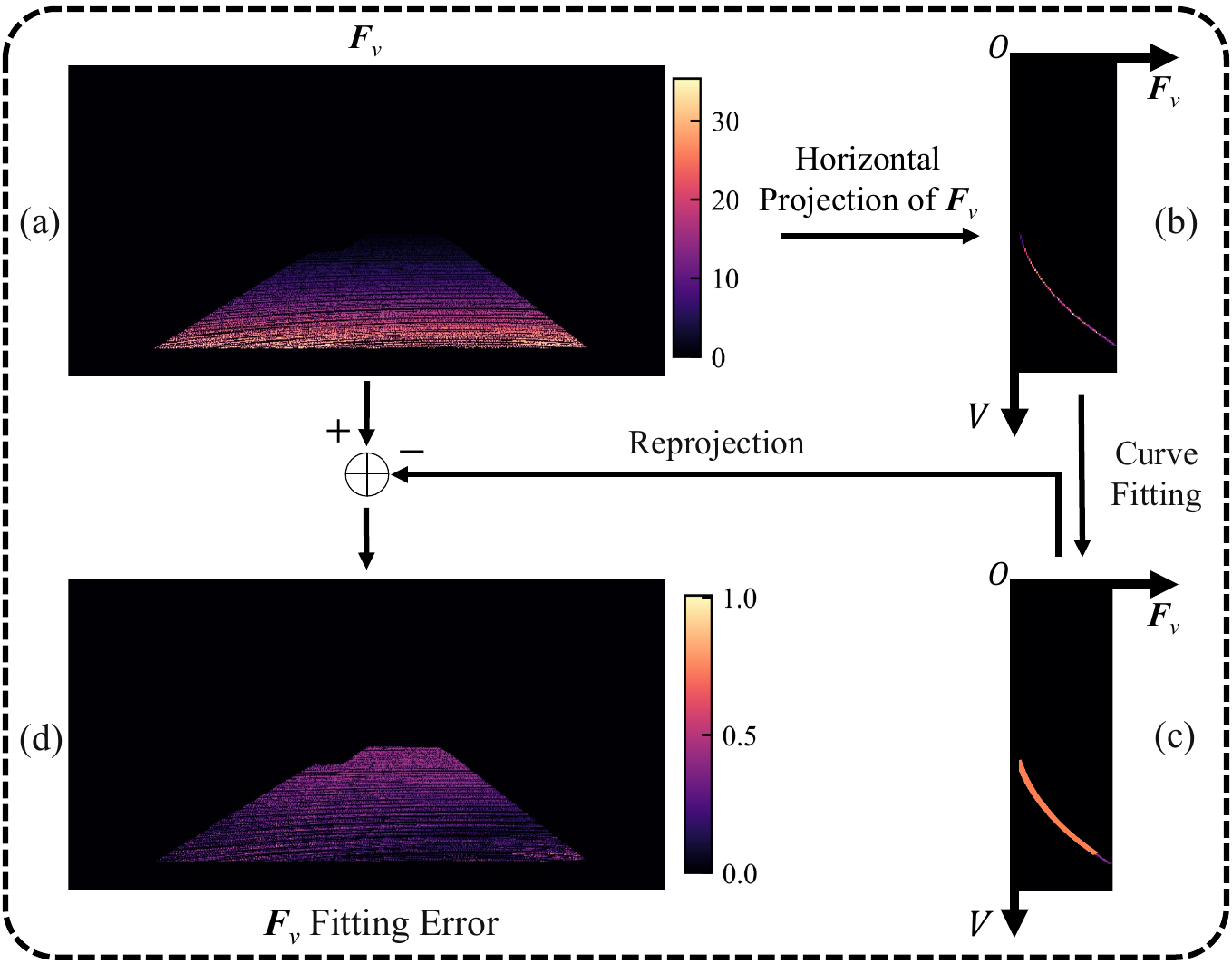}
			\caption{Optical flow modeling in the special case: (a) $\boldsymbol{F}_v$; (b) horizontal projections of $\boldsymbol{F}_v$; (c) curve fitting result; (d) $\boldsymbol{F}_v$ fitting error visualization.}
			\label{fig.curvefit}
		\end{figure}

            We also conduct an additional experiment to demonstrate the robustness of our optical flow modeling process in the special case where $\theta=0$, $\varphi=0$, and $x_d=0$. As shown in Fig. \ref{fig.curvefit}, similar to the V-Disparity analysis process, we create the histogram of $\boldsymbol{F}_v$ for each row and then fit the $\boldsymbol{F}_v$ projections to a curve expressed in (\ref{eq.fdfv_simple}). A fitted $\boldsymbol{F}_v$ map can then be generated using the curve parameters. By comparing the difference between the observed and fitted $\boldsymbol{F}_v$ maps, we obtain an optical flow modeling error of less than one pixel, providing compelling evidence for the validity of our model in this special case.

             \begin{figure}[!t]
			\centering
			\includegraphics[width=0.48\textwidth]{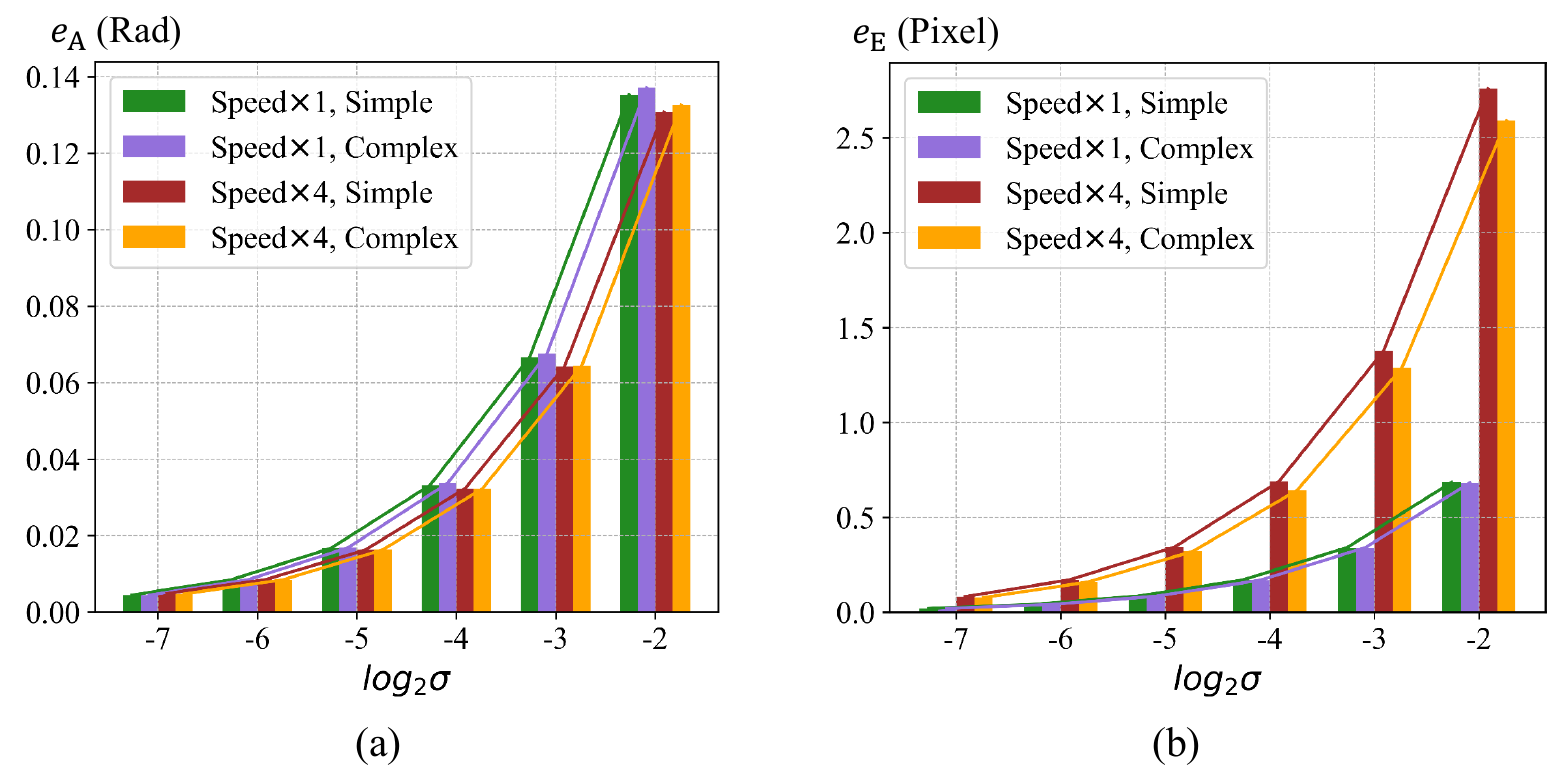}
			\caption{Noise robustness experiment on the CVC12 dataset. $\sigma$ represents the variance of added Gaussian noise.}
			\label{fig.noise}
		\end{figure}

		  \begin{table}[t!]
  \fontsize{7.6}{11.5}\selectfont
			\begin{center}
				\caption{
					Comparison of the photometric error achieved by two state-of-the-art optical flow estimation networks w/o and w/ our proposed algorithm incorporated.}
				\label{table.platform}
				{
					\begin{tabular}{c|cc|cc}
						\toprule
						
						\multirow{2}{*}{Networks}	
						& \multicolumn{2}{c|}{Indoor} 	
						& \multicolumn{2}{c}{Outdoor}    \\
						\cline{2-5}
						
						& Baseline   & Our model    &Baseline    &Our model \\
						
						\hline
						
						RAFT \cite{teed2020raft}   &10.423	&8.266	&9.252 &7.399\\
						ARFlow \cite{liu2020learning}	&11.251	&9.125	&9.953 &8.017 \\
						
						\bottomrule
					\end{tabular}
				}
			\end{center}
		\end{table}
  
		\subsection{Robustness to Random Noise}

            To further evaluate the robustness of our proposed model, we conduct an experiment where random Gaussian noises with varying intensities are added to the CVC12 dataset. We then compare the fitting results with the observed optical flow ground truth. As depicted in Fig. \ref{fig.noise}, our model demonstrates remarkable robustness, effectively ignoring noise even when the variance $\sigma$ is less than $1/32$. Additionally, our proposed model exhibits superior performance in terms of end-point error at lower vehicle speeds and maintains consistency across different scenarios. Furthermore, the model demonstrates robustness in terms of the average angular error across various scenarios and vehicle speeds.

        \subsection{Real-time Performance}
        
We conduct an experiment to evaluate the efficiency of our proposed model on an Intel i7-12700K CPU (using only a single thread). With the input of odometry and camera parameters, our model can generate accurate freespace optical flow (resolution: 1242$\times$375 pixels) at a speed of 34.5 frames per second, without any hardware acceleration.

		\subsection{Practical Experiments}
		\label{sec.practical_experiments}
		
		\begin{figure}[!t]
			\centering
			\includegraphics[width=0.49\textwidth]{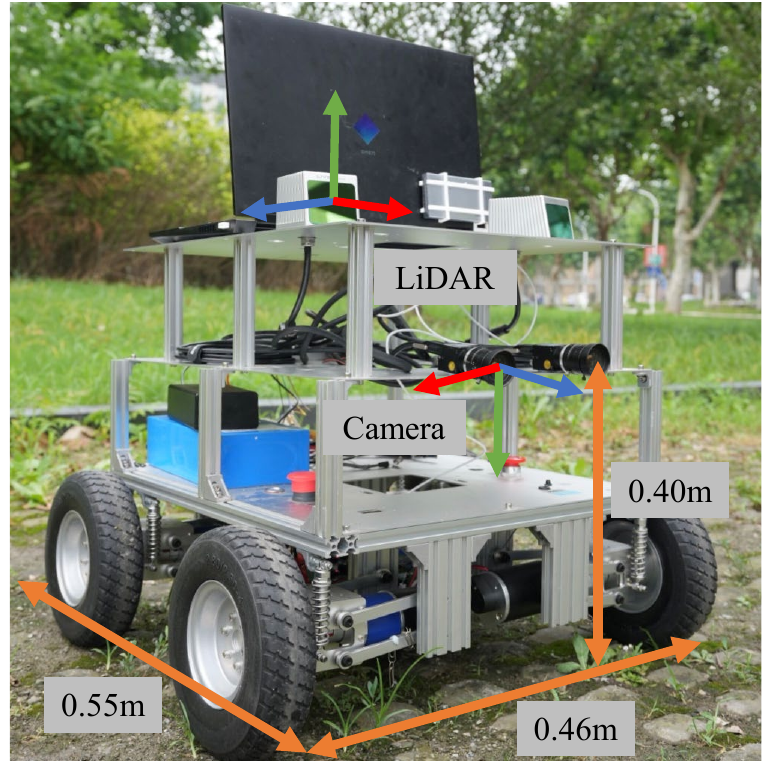}
			\caption{An illustration of our practical experiments, where all sensors are well-calibrated and synchronized to ensure accurate data collection.}
			\label{fig.platform}
		\end{figure}

		\begin{figure}[!t]
			\centering
			\includegraphics[width=0.49\textwidth]{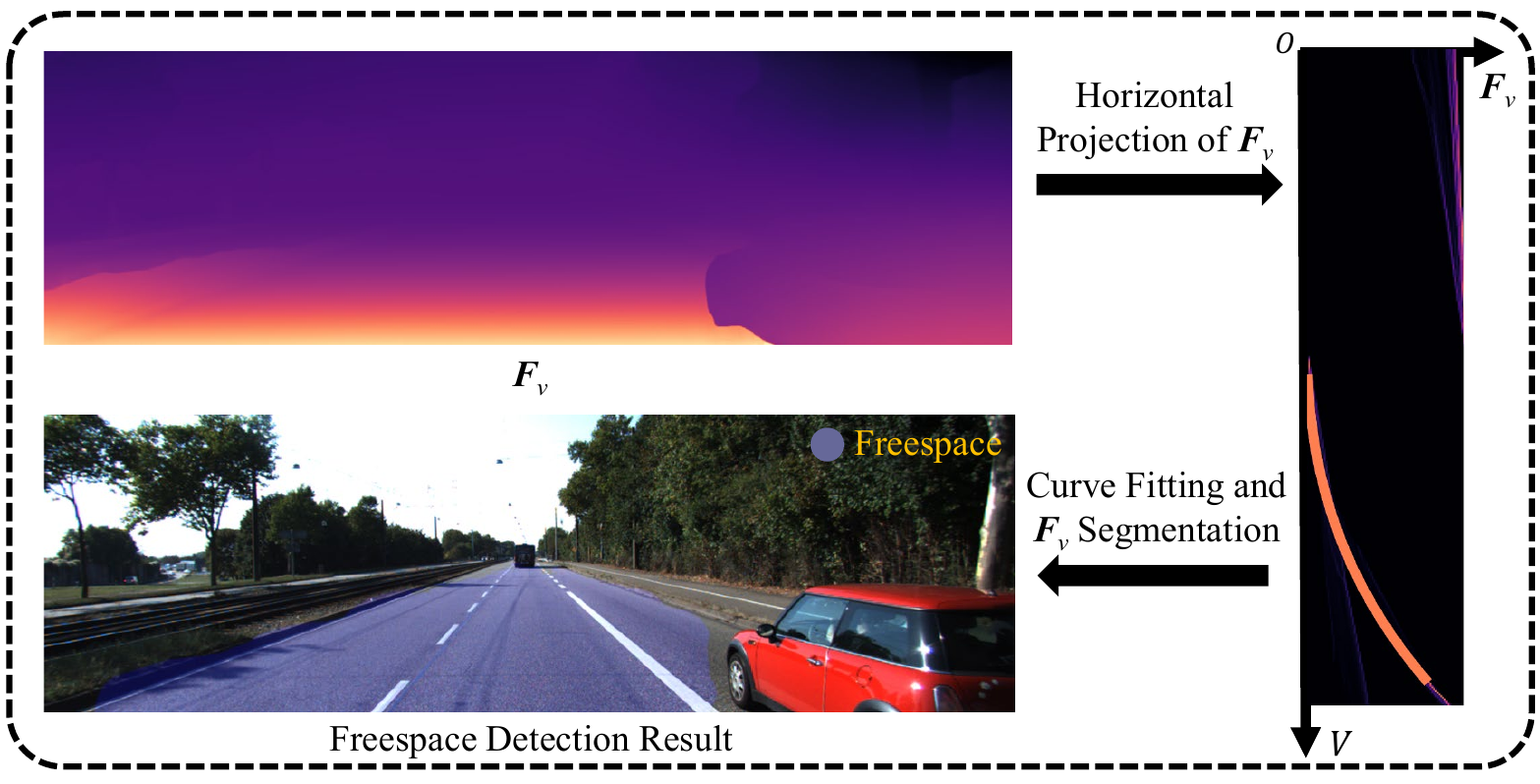}
			\caption{Freespace detection using our proposed displacement-based optical flow model.}
			\label{fig.fs1_seg}
		\end{figure}

		To validate the feasibility and effectiveness of our proposed freespace optical flow modeling approach, we conduct extensive practical experiments on an automated guided vehicle, as illustrated in Fig. \ref{fig.platform}. We utilize a well-calibrated stereo rig to collect video sequences in both indoor and outdoor environments. The optical flow information was obtained using pre-trained optical flow networks RAFT \cite{teed2020raft} (supervised) and ARFlow \cite{liu2020learning} (unsupervised). We compare the optical flow estimation results without and with our proposed algorithm incorporated. We apply the optical flow to warp the target image to the reference view and calculate the photometric error between the warped target image and the reference image. The results reported in Table \ref{table.platform} indicate that our proposed displacement-based model can effectively improve the freespace optical flow results obtained using other deep learning-based algorithms. By applying our modeling approach as a post-processing step, we are able to improve the geometry consistency of the freespace optical flow. These findings highlight the potential of our approach in enhancing the performance of existing optical flow algorithms for autonomous driving perception and navigation tasks.

        \begin{figure}[!t]
			\centering
			\includegraphics[width=0.49\textwidth]{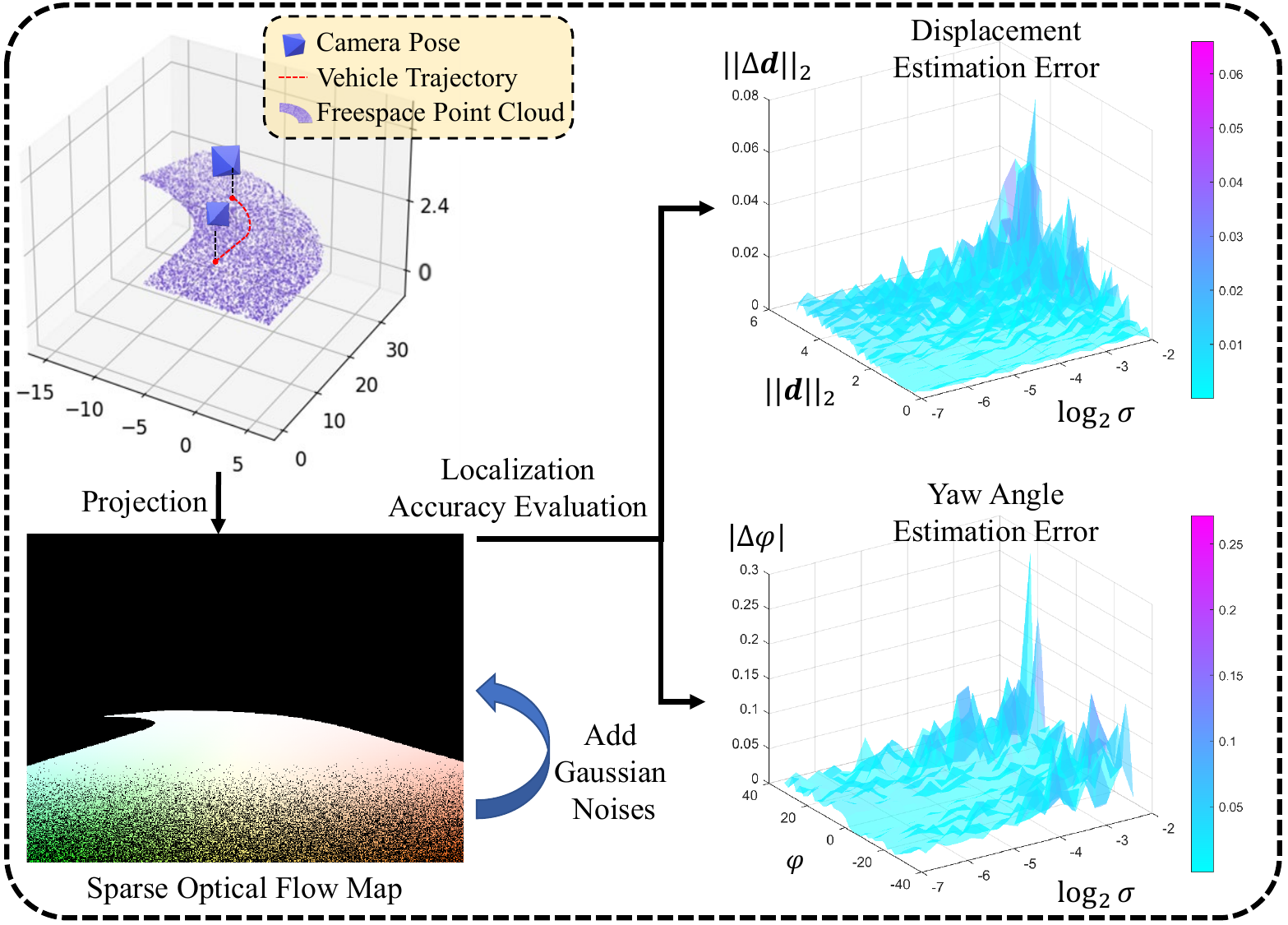}
			\caption{Vehicle pose estimation with respect to varying displacements, yaw angles, and different levels of Gaussian noise.}
			\label{fig.localization}
		\end{figure}

		\section{Potential Autonomous Driving Applications}
		\label{sec.discussion}		
		
		\subsection{Freespace Detection}
         
            As discussed in Section \ref{sec.optical_flow_special_cases}, we can leverage our proposed displacement-based optical flow model to segment $\boldsymbol{F}_v$ for freespace detection. Fig. \ref{fig.fs1_seg} shows an example of freespace detection results achieved using the technique depicted in Fig. \ref{fig.curvefit}. This process is akin to the road area detection with disparity segmentation presented in \cite{fan2019road}.
  
		\subsection{Vehicle Pose Estimation}
            Our proposed model also holds great promise for vehicle pose estimation by utilizing the information provided by freespace optical flow. As depicted in Fig. \ref{fig.localization}, we begin by creating a virtual road plane using uniform sampling points. Assuming that the vehicle has a random displacement and yaw angle changes while moving along this virtual road plane, we employ the pinhole camera model to obtain the optical flow ground truth. To evaluate the robustness of our model against measurement noise, we introduce Gaussian noise with varying variances $\sigma$ to the optical flow ground truth, which serves as the sole input to our pose estimation pipeline. We employ the Particle swarm optimization algorithm \cite{wang2018particle} in the bivariate fitting process to determine the best localization parameters. The absolute estimation errors of $\boldsymbol{d}$ and $\varphi$ are shown on the right. It is evident that the displacement estimation error is below 0.07m, and the yaw angle estimation error is less than 0.3$^\circ$. Moreover, the localization accuracy demonstrates robustness against noises of different magnitudes of displacement and rotation, indicating the effectiveness of our approach in accurate pose estimation even under challenging conditions.

		\section{Conclusion}
		\label{sec.conclusion}
		This article presented two freespace optical flow models for automated driving based on two definitions (displacement and velocity). By leveraging the monocular camera parameters and vehicle poses, we established an explicit relationship between optical flow and the pixel coordinates. We mathematically demonstrated the distribution regularities of both optical flow models in the pixel coordinate system. To verify the validity and accuracy of our proposed models, we conducted extensive experiments on the CVC12 optical flow dataset, KITTI flow 2012 and 2015 datasets, and the CARLA dataset. The average end-point error is less than one pixel on the CVC12 and KITTI datasets and less than 1.5 pixels per second on the CARLA dataset. We also verified the validity of our model in special cases via curve fitting. Our proposed model performs well against noise and demonstrates real-time performance. Finally,  we showcased the potential applications of our model through experiments involving freespace detection and vehicle pose estimation.

		\bibliographystyle{IEEEtran}


	\end{document}